# Impact of Argument Type and Concerns in Argumentation with a Chatbot

Lisa A. CHALAGUINE [a] Fiona L. HAMILTON [b] Anthony HUNTER [a]
Henry W. W. POTTS [c]

[a] *Computer Science Department, University College London, London, UK*
[b] *eHealth Unit, University College London, London, UK*
[c] *Institute of Health Informatics, University College London, London UK*

**Abstract.** Conversational agents, also known as *chatbots*, are versatile tools that have the potential of being used in dialogical argumentation. They could possibly be deployed in tasks such as persuasion for behaviour change (e.g. persuading people to eat more fruit, to take regular exercise, etc.) However, to achieve this, there is a need to develop methods for acquiring appropriate arguments and counterargument that reflect both sides of the discussion. For instance, to persuade someone to do regular exercise, the chatbot needs to know counterarguments that the user might have for not doing exercise. To address this need, we present methods for acquiring arguments and counterarguments, and importantly, meta-level information that can be useful for deciding when arguments can be used during an argumentation dialogue. We evaluate these methods in studies with participants and show how harnessing these methods in a chatbot can make it more persuasive.

## 1. Introduction

Chatbots are versatile tools that have the potential of being used as agents in dialogical argumentation systems for behaviour change applications. For example, a chatbot could persuade people to do more sports by presenting arguments in favour of exercising and countering the arguments given by the user as to why she is not willing to. The chatbot is thereby engaging in an argumentation dialogue where it acts as the persuader and the user as the persuadee. This calls for the development of methods for acquiring appropriate arguments and counterarguments that reflect the points of view of both parties. In the example above, this could include arguments why exercise is healthy, as well as arguments a user might give for not doing regular exercise. The chatbot needs to be aware of the potential arguments people might have for not engaging in the behaviour in question, in order to reply with appropriate counterarguments. However, there can be many counterarguments to choose from, and their degree of impact can vary. Therefore, meta-level information about the arguments could help the chatbot in using them effectively.

In persuasion, the way an argument is communicated is just as important as its message. A persuader who wants to convince a persuadee to do more exercise can present his argument in many different ways. He can, for example, point out the advantages of regular exercise: *"Regular exercise will strengthen your bones, muscles, and joints"*. However, he could also phrase it in a negative way: *"Lack of regular exercise leads to weakening of your bones, muscles, and joints"*. This notion of *framing* is well studied in psychology and health care [16,17,22,15]. Other persuasion techniques such as referral to authority and social proof [8] have also been used in psychology. We refer to the style of persuasion used in an argument, as argument *type*.

We want to investigate some common argument types used in persuasive dialogues in the behaviour change domain. Despite the extensive psychology literature on the topic of message framing and persuasion techniques, the notion of argument type is underdeveloped in the computational argumentation field. Walton's *argumentation schemes* [24] could be viewed as a non-exhaustive summary of argument types. However, that leaves some important types for behaviour change unconsidered. Also, to the best of our knowledge, no empirical studies with participants were undertaken to test the effectiveness of argument types in persuasion.

Furthermore, we want to investigate how *concerns* of the persuadee impact the effectiveness of arguments in persuasion dialogues. The results in [12] show that taking persuadee's concerns into account improves the persuasiveness of a dialogue. The persuader might present a valid argument the persuadee does not disagree with (e.g. that regular exercise is good for her health), but which has no impact on the persuadee because she is not concerned with her health at that moment. However, she might be very concerned with her academic performance. The argument *"Regular exercise not only help your physical health, but it will help you study better"* might, therefore, have a bigger impact on the persuadee because it addresses her concern.

In this paper we investigate different *types* of counterarguments and their preference with the persuadee based on the persuadees *concerns*. We propose a method for crowdsourcing arguments and counterarguments and assess a typology of counterarguments and concern assignments to be used by a chatbot. We use meat consumption as a case study. To verify our approach, we developed a strategic chatbot that takes the concern of the user into account and during an argumentation dialogue with the user, presents only those types of counterarguments that address his or her concern. For comparison purposes, we also developed a baseline chatbot that does not address the user's concerns. Our results show that the strategic chatbot outperforms the baseline one, and has a more significant impact on the user's intention to reduce their meat consumption in the future.

This research builds upon our previous work [6] in which we explored *argument harvesting*, i.e. methods for acquiring arguments with the help of a chatbot. Our results showed that a chatbot with no or limited domain-specific knowledge can acquire meaningful arguments. That was achieved through the use of a particular query system, thus making the chatbot act only as an inquirer. In contrast, in this paper, the chatbot can engage in a fully fledged argumentation dialogue with the user and thus act as a persuader.

The rest of the paper is structured as follows: Section 2 gives some background theory on the notions of *concerns* and *appeal*, different types of counterarguments, and presents the ones we are investigating in this paper; Section 3 presents the aim of the paper and the hypotheses; Section 4 describes the argument and counterargument acquisition process; Section 5 describes the experiments that were conducted with the acquired data, namely the evaluation of the counterargument types and the chatbot that was used for the persuasion dialogue; and in Section 6 we discuss and conclude our findings.

## 2. Conceptualising Argumentation

**Arguments:** In our study we chose the topic *meat consumption* and were interested in different argument types in favour of reducing meat consumption which we could present to meat eaters as counterarguments to their arguments in favour of eating meat. It is important to point out that we acquired our arguments by crowdsourcing. We opted for this method because there exists no central repository of all possible counterarguments on a certain topic. Depending on the topic, there might be information available and possibly even a summary of arguments in the professional literature, e.g. why reducing meat consumption is good for you and/or the planet. However, what might convince a professional, might not necessarily convince the average meat eater. In addition, crowdsourcing offers a fast and efficient way to gather a large number of diverse arguments without introducing the researcher's bias into the selection of arguments if gathering them by hand. It should be noted that the same could be done for any other topic.

**Argument Types:** A persuader who wants to convince a persuadee to do more exercise can present his argument in many different ways. We refer to the style of persuasion used in an argument, as argument *type*. We investigate six argument types in total. The most common type of counterargument used in the computational argumentation literature is the *negation* of an argument. We, therefore, included a type which we call *direct counter-*

**Table 1.** Definitions of six investigated argument types with examples. The argument countered in the examples is *"I eat meat because it tastes good."*.

| Argument Type | Definition | Example |
|---|---|---|
| Direct Counter-argument | This argument is counterargument that directly negates a previously given argument by referring to it. | *A raw, unprepared chunk of meat doesn't taste good. It's about the way of preparation and seasoning.* |
| Suggestion-based Argument | This argument gives a suggestion that may implicitly refer to a previously given argument and suggests how to change the behaviour in question | *You could introduce one day a week where you don't eat meat and try different substitutes instead. Overtime you can increase the number of days.* |
| Positive Personal Consequence | This argument gives a negative consequence for the persuadee personally, if he or she continues the behaviour in question. | *Eating less meat will decrease your cholesterol level which will ultimately lower your risk of stroke and heart disease.* |
| Positive Impersonal Consequence | This argument gives a positive consequence for someone/something apart from the persuadee, if he or she continues the behaviour in question. | *Eating less meat will lead to the reduction of the water foot print on the earth.* |
| Negative Personal Consequence | This argument gives a negative consequence for the persuadee personally, if he or she continues the behaviour in question. | *Most processed meats are loaded with artificial chemicals, including flavourants, colourants and preservatives that might be bad for your body.* |
| Negative Impersonal Consequence | This argument gives a negative consequence for someone/something apart from the persuadee, if he or she continues the behaviour in question. | *Much land is needed to raise cattle for which forests have to be cut down, therefore causing deforestation.* |

*argument* which negates the given argument. In our previous study [6], when people were asked to provide a counterargument to their given argument, people mostly gave arguments in the form of suggestions. Suggestions are often enthymemes and do not explicitly negate the argument. They, however, imply that changing the behaviour is advantageous (therefore attacking the argument) and provide a solution on how to achieve that. For this reason, we include suggestions, or *suggestion based arguments* in our argument types assessment.

As mentioned above, arguments can be framed in a positive or negative way, either referring to a gain or a loss (i.e. positive or negative consequence for the persuadee). We, therefore, included *positive* and *negative consequences* into our list of argument types we want to explore. Further, a certain behaviour often has consequences not just for the person engaging in that behaviour but on others as well, which we call *personal* and *impersonal* consequences respectively. Smoking, for example, is not just bad for the smoker but also imposes a burden on the health care system if the smoker potentially becomes sick due to his behaviour. This way, we end up with six argument types we want to investigate. We give the definition and an example for each type in Table1.

**Concerns:** Arguments can raise or address various concerns of the persuadee that need to be accounted for. A persuader might present a perfectly valid argument, e.g. *"Meat consumption has a negative impact on the environment as it causes deforestation as huge tracts of rain forest are burned for pasture"*. The persuadee might not even disagree with this argument, however, if she is not concerned about the environment, this argument may have no impact on her intention to change her behaviour. If, however, the persuadee is concerned about her health, then the argument *"Some meats are high in saturated fat. Eating a lot of saturated fat can raise cholesterol levels, which raises your risk of heart disease"* is more likely to change her intention to consume less meat. Whilst this is a simple and intuitive idea, there is a lack of a general framework for using concerns in making strategic choices of move in the way suggested by the above example [12]. A similar notion to concerns are *values*. In value-based argumentation [3], values are assigned to an argument when constructing argument graphs. They provide an explanation as to why it is not always possible to persuade others to accept an opinion simply by demonstrating facts and proofs. It may be that a particular individual will accept the facts of a decision but will reject the conclusion to act upon it because it does not support the values he or she holds [2]. In value-based argumentation frameworks (VAFs), values are used for ignoring attacks by counterarguments where the value of the attacking argument is lower ranked than the attacked one. Thus, the role of values in VAFs is different from the role of concerns as suggested in the example above where the persuader chooses the argument of greater concern to the persuadee [12].

From preliminary investigation which involved researching the most common arguments against meat consumption on the internet, we discovered that most arguments revolve around two major *concerns*: *Health* and *Environment*. Note, we can view the health concern as a *personal* concern (i.e. something that directly relates to the individual), and we can view the environment concern as an *impersonal* concern (i.e. something that does not directly relate to the individual).

**Appeal:** In this study we are interested in the appeal of the argument *type*, not the argument itself. As pointed out by [18] an appealing argument might not necessarily be a convincing one. The argument that education should be free might be very appealing, but at the same time, we can acknowledge that universities need resources to function and therefore not be very convincing. The same is true for the contrary: as unappealing as the argument for immediate treatment of a brain tumour might be, the patient will probably be convinced to undergo the operation. We, however, are interested in the appeal of the *type* of argument and not in the appeal or convincingness of the *message* of the argument. We believe that one type of argument cannot be more convincing than another type, but one type can indeed be more appealing and more in line with the concerns of the persuadee than another. For example, given two arguments with a message of "equal convincingness" (however one chooses to measure that), one argument addressing health benefits and the other argument addressing the environmental benefits of reducing meat consumption, a persuadee might find the argument addressing health more appealing to her, because she is not concerned about the environment as much as she is concerned about her health.

## 3. Hypotheses

In this paper, we show how the persuader's choice of argument type and concern influences the persuadee's intention to change his or her behaviour. Firstly, we explore different *types* of counterarguments and evaluate their appeal to the participants in the behaviour change domain. Secondly, we investigate whether the persuadee's concerns have an impact on the argument types the persuadee found most appealing. Thirdly, we use a chatbot in order to test whether presenting only those counterarguments that address the persuadee's concern is more likely to change his or her intention to the positive than presenting counterarguments that address other concerns in the domain as well.

We summarise these points in the following three hypotheses:

**H1** When a person is presented with counterarguments of various types, some types are perceived as more appealing than others.

**H2** When a person is presented with counterarguments that address different concerns in that domain, people find those counterarguments more appealing that address the concern that they perceive as more important.

**H3** A chatbot with no natural language understanding, just by presenting the type of counterarguments that take the user's concern into consideration, is more likely to have a positive impact on changing the user's attitude, than a chatbot that presents the type of counterarguments that ignore the user's concern.

In the remainder of this paper we describe the methods for acquiring the main arguments why people eat meat, different types of counterarguments for meat consumption and explain the experiments conducted with them in order to test our hypotheses.

## 4. Argument & Counterargument Acquisition

Our study consisted of two parts: the argument and counterargument acquisition, described in this section, and the experiments (described in the next section) which we conducted with the acquired data in order to test our hypotheses.

The participants for all surveys and experiments were recruited via *Prolific*[1], which is an online recruiting platform for scientific research studies. For each survey, we recruited from either one or two disjoint groups: meat eaters and vegetarians. We opted for this division in order to evaluate counterarguments obtained from people who engage in the behaviour in question (in this case, meat consumption) and people with the opposite behaviour (abstaining from eating meat). The general prerequisites for taking part in our study were to be over 18 and fluent in the English language. We used Google Forms for all surveys.

The aim of this part of our study, the crowdsourcing of arguments and counterarguments, was fourfold: (1) To identify the most popular arguments for eating meat amongst the participants; (2) To acquire counterarguments for the most popular arguments from meat eaters and vegetarians (*direct* counterarguments); (3) To evaluate whether meat eaters preferred the counterarguments given by fellow meat eaters or by vegetarians; (4) And, given the results from (3), to acquire the remaining five types of counterarguments from the group whose *direct* counterarguments were preferred by the meat eaters.

*4.1. Step A1: Argument Acquisition and Clustering*

In order to find the most popular arguments for eating meat amongst the participants, we recruited 40 meat eaters and asked them in a Google Form to give their main reasons for eating meat. This way we acquired 111 arguments which can be found in Appendix I [1]. The average length of an argument was 7 words with a standard deviation of 5.

We used the algorithm described in [6] to automatically preprocess and cluster the arguments. Arguments were clustered by similarity, in order to identify the most popular arguments for eating meat. We clustered 81 arguments (73%) into 7 clusters. However, we merged the cluster where the majority of arguments were *"I like it"* with the cluster where the majority of arguments were *"It tastes good"* because we decided that those are in fact the same argument (people like eating meat because of its taste). The remaining clusters we left the way our algorithms clustered them. The 30 arguments that could not be clustered were either the only one of its kind or despite saying similar things, were too differently worded, e.g. *"Because I was brought up eating meat"* and *"Because our parents always gave us meat"*. We did not use the unclustered arguments. As a representative argument for each cluster, we randomly picked one of those that contained the highest number of *most common words* in that cluster. In the rest of the paper, those will be referred to as "the most popular arguments" for eating meat. The size of the clusters and the representative argument of each are given in Table 2. The name of each cluster is the most common word found in that cluster (excluding stopwords).

*4.2. Step A2: Direct Counterargument Acquisition for Most Popular Arguments*

After identifying the most popular arguments for eating meat, we started with the direct counterargument acquisition. This consisted of two steps:
**Step A2i.** For each of the most popular arguments obtained in step A1, we created two surveys. For the first, we recruited 10 meat eaters, and for the second 10 vegetarians and asked them to counter the given argument by giving a single argument. This way we ac-quired 20 direct counterarguments for each of the most popular arguments: 10 given by meat eaters and 10 given by vegetarians.

---
[1] www.prolific.ac

Table 2. Summary of the arguments acquired in Step 1. Cluster name, number of arguments in that cluster and representative argument for that cluster.

| Cluster | No of Args | Representative Argument |
|---|---|---|
| Nutrition | 15 | *For its nutritional value and source of protein* |
| Filling | 6 | *It's filling* |
| Taste/Like | 40 | *It tastes good* |
| Easy | 6 | *Quick and easy to prepare* |
| Health | 11 | *It's healthy and contributes to a balanced diet* |
| Variety | 3 | *It offers more variety to my meals* |

**Step A2ii.** We were interested only in the "best" counterarguments given in step A2i and wanted to identify whether meat eaters or vegetarians give the more effective counterarguments. We, therefore, created two surveys (one for meat eaters and one for vegetarians) and recruited 20 participants for each survey. For each argument obtained in step A1, the participants were presented the argument and the 20 acquired counterarguments. Since we were not interested in the message of the counterargument (e.g. its believability or convincingness) but still wanted clear, understandable and appropriate representatives of each counterargument type, we asked the participants to select those counterarguments that they found best at communicating their message. We counted the number of times each counterargument was voted for and ordered them by the number of votes. It should be noted that we did not determine the truthfulness of the arguments. It is, therefore, possible that the message of some counterarguments is not factually correct.

The vegetarian participants on average selected 9.7 out of 20 counterarguments, while the meat eaters only selected 7.55. This is not surprising, as the vegetarians who do not eat meat anyway, are likely to be less skeptical about the counterarguments. Interestingly, both groups found the counterarguments given by meat eaters slightly more appealing in total. Vegetarians on average selected 5.1 out of the 10 counterarguments given by meat eaters and only 4.6 given by other fellow vegetarians. Meat eaters on average selected 3.95 counterarguments given by other meat eaters and only 3.6 by vegetarians. The highest ranked counterargument for each argument, however, was given by a vegetarian and for 5 out of the 6 most popular arguments, the two highest ranked counterarguments were given by a vegetarian. We, therefore, opted for the vegetarians as the source of the remaining counterarguments types. The data for this survey is summarised in Table 3 and all counterarguments can be found in Appendix II a [1].

### 4.3. Step A3: Non-direct Counterargument Acquisition

We acquired the remaining types of counterarguments from vegetarians only, as the highest ranked counterarguments were given by vegetarians during the direct argument acquisition described in the previous section.

Table 3 Counterarguments (CA) for eating meat given by meat eaters (ME) and vegetarians (V) and the average number of times they were selected by meat eaters and vegetarians for the six most popular arguments for eating meat.

| CA given by | Selected by ME | Selected by V |
|---|---|---|
| ME | 79 | 102 |
| V | 72 | 92 |

**Step A3i.** We created one survey for which we recruited 10 vegetarians and asked them to provide one suggestion on how to reduce meat consumption, one negative personal consequence and one negative impersonal consequence for continuing meat consumption, and one positive personal consequence and one positive impersonal consequence for reducing meat consumption. They were given the same examples of these argument types as given in Table 1.

**Step A3ii.** Using the same approach as in A2i, we again received a ranking which allowed us to identify the best counterarguments (those ranked the highest) and allowed us to make comparisons of the voting between meat eaters and vegetarians.

*4.4. Results of Steps A2 and A3*

Table 4 shows the total votes by vegetarians and meat eaters for the different argument types. There were 10 counterarguments and 20 participants per argument type. Therefore, the maximum number of votes an argument type could acquire was 200 (had all arguments been selected by all participants). It can be seen that vegetarians again selected more counterarguments to all argument types. Both groups selected *Negative Impersonal Consequences* the most, whereas *Direct Arguments* were least selected by both groups.

## 5. Experiments

The experiments were split into two parts: the first was concerned with the evaluation of the different argument types according to their appeal, and their concern assignment. We wanted to test whether the concerns *health* and *environment* for the (im)personal, positive and negative consequences do in fact represent the concerns of the participants. In the second part, we used a chatbot in order to test whether presenting counterarguments that take the user's concerns into account was more likely to change the user's attitude to the positive compared to a chatbot presenting counterarguments that ignore the user's concerns.

As the chatbot was used to chat with meat eaters, we used the three counterarguments that were ranked the highest by meat eaters in the steps A2ii and A3ii for each counterargument type. This resulted in a total of 18 direct counterarguments (three for each of the six most popular arguments for eating meat) and the top three ranked of the remaining five types. These counterarguments can be found in Appendix II b [1].

**Table 4** Summary of counterarguments acquired in step A3. Total votes by the 20 vegetarians and 20 meat eaters for the different argument types. For *DIR* the number of votes divided by two since there were 20 counterarguments to choose from in step A2ii, as opposed to 10 in step A3ii. *DIR = Direct Argument, SUG = Suggestion, NIC = Negative Impersonal Consequence, NPC = Negative Personal Consequence, PIC = Positive Impersonal Consequence, PPC = Positive Personal Consequence.*

| Meat Eaters | | Vegetarians | |
|---|---|---|---|
| **Arg Type** | **Votes** | **Arg Type** | **Votes** |
| NIC | 97 | NIC | 136 |
| PIC | 91 | PIC | 133 |
| SUG | 91 | NPC | 115 |
| NPC | 90 | PPC | 112 |
| PPC | 76 | SUG | 106 |
| DIR | 75 | DIR | 97 |

## 5.1. Evaluation of Argument Types

In this part of the experiment, we evaluate the six different argument types according to their appeal to the participants and show the correlation between preferred argument type and the concerns of the participants in order to investigate hypothesis H1 and H2.

### 5.1.1. Methods

We created a survey that consisted of five parts with the following questions. **(S1)**: the participants were asked to provide their age, gender, education, occupation and number of children. **(S2)**: they were asked about their frequency of meat consumption (choice of 1-2 times per week/3-4 times per week/5-6 times per week/every day/several times a day. **(S3)**: they were asked what their main reason for eating meat was. There was a choice of six as shown in Table 2 and the option "other". **(S4)**: they were presented with the 3 highest ranked counterarguments of each type. They were asked to pick all the counterarguments that appeal to them. Note, that if they selected the option "other" in the previous step, no direct counterarguments were presented. **(S5)**: they were asked to provide a short explanation of why they chose those counterarguments. 100 meat eaters were recruited for this survey.

### 5.1.2. Results

In this section, we only present the results relevant to the development of the chatbot. The whole data can be found in Appendix III [1].

We were interested in two things: Firstly, whether there is a difference in popularity of argument types. And secondly, whether there is a correlation between the preferred argument type of the participants and any of the information that they provided which the chatbot could take into consideration when presenting the arguments during an argumentation dialogue. Table 5 shows the normalised data from the meat eaters in Table 4 (left) obtained in step A3ii and the data from this experiment (right). The percentages show how many of the shown argument types were selected overall by all participants, i.e. there were 3 arguments of each type, since 33% of the *Suggestion-based arguments* (SUG) were selected it means that on average each participant selected at least one SUG. It is interesting to contrast these results with the results we got in step A3ii where only one argument type was presented per survey, whereas during this experiment, arguments of each type were presented together and therefore put in contrast to each other. One can clearly see that *Direct Counterarguments* are much less popular compared to the others. The four types of consequential arguments were the most popular, and *Negative Personal Consequence* ranked the highest. The results support our H1, that different types of counterarguments differ in their appeal.

Next, we wanted to find out on what user attributes the preference of counterargument type depends on. This is valuable because future chatbots could use this to get this information from the participant before presenting the tailored counterarguments dependent on that attribute. No correlation was found between the preferred argument type of the participant and any of the information provided apart from their explanations in (S5). We observed that the explanations of most participants raised concerns about their health or the environment, or both, which further supports our choice of concerns for this do-main. We therefore automatically assigned concerns to the explanations. Every explanation that contained the word *health* was assigned the concern *Health* and those that contained the words *animal, environment, planet* were assigned the concern *Environment*. Explanations that contained words from both concerns, were labeled *Both*.

Table 5 Comparison of ranking obtained in step A3ii and the ranking obtained in the first part of the experiment. See caption for Table 4 for acronyms.

| Ranking | | Experiment | |
|---|---|---|---|
| Arg Type | Selected | Arg Type | Selected |
| NIC | 49% | NPC | 55% |
| PIC | 46% | PPC | 51% |
| SUG | 45% | NIC | 50% |
| NPC | 45% | PIC | 47% |
| PPC | 38% | SUG | 33% |
| DIR | 38% | DIR | 6% |

We observed a high statistical correlation between the participants' concerns and their preferred argument type. Participants who gave an explanation that was labeled with the concern *Health* preferred the positive and negative *personal* consequences, whereas participants that gave an explanation that was labeled with the concern *Environment* preferred the positive and negative *impersonal* consequences. Participants who gave an explanation that was labeled with both concerns preferred all consequential counterarguments equally. We used the Chi-Square test in order to calculate statistical significance by comparing the numbers of the available counterarguments with the number of the selected ones. For example, since 28 participants were concerned about health only, there were 168 (28 x 6) personal consequential arguments to select from and 336 (28 x 12) of the remaining types. 120/168 out of the personal consequential arguments were selected in contrast to only 70/336 of the remaining types. The p-values for all three groups were below 0.001. The results are summarised in Table 6. The results support our hypothesis H2, that people strongly prefer argument types that relate to their concern.

### 5.2. Chatbot

We developed two versions of the chatbot, one that took the concern of the user into account when presenting counterarguments (strategic chatbot), and one that did not (baseline chatbot). The purpose of the chatbot was twofold: firstly, to test whether presenting counterarguments that take the user's concern into consideration is more likely to change the user's attitude to the positive, than presenting counterarguments that ignore the user's concern. Secondly, to test whether a chatbot, that has no natural language understanding can engage in an argumentation dialogue, harvest arguments and influence the user's attitude about the discussed topic. With natural language understanding we mean, that the chatbot does not "understand" what the user writes, i.e. no keyword matching or machine learning. So this experiment was to investigate hypothesis H3.

Table 6 Summary of the results of the first part of the experiment. Total number of selected counterarguments per concern. See caption for Table 3 for acronyms.

| | Argument Type | | | | | | |
|---|---|---|---|---|---|---|---|
| Concern | DIRECT | SUG | NIC | NPC | PIC | PPC | No of participants |
| Health | 7 | 26 | 22 | **65** | 15 | **55** | 28 |
| Environment | 2 | 25 | **59** | 31 | **59** | 22 | 31 |
| Both | 3 | 17 | **39** | **40** | **39** | **42** | 18 |

*5.2.1. Methods*

The chatbot was deployed on Facebook via the Messenger Send/Receive API. For more on the implementation of such a chatbot see [5]. For each chatbot we recruited 50 participants. The dialogue protocol, described in dialogue steps DS1 to DS9 was as follows:

**DS1** The participant was asked at the beginning of the chat to select whether they would consider reducing their meat consumption. The choices were: *definitely wouldn't, probably wouldn't, might, probably would* and *definitely would*.

**DS2** Then they were asked what they were more concerned about: the impact that meat consumption had on their health or the impact it had on the environment and animals. They were given two options to select: *health* and *environment/animals*.

**DS3** Then they were asked to select their main argument for eating meat (see Table 2) and the option "other", to start the chat.

**DS4** Then the chatbot presented its first counterargument. For the chatbot we only used the four consequential types of counterarguments, since they scored the highest during the first part of the study, described in section 4.

**DS5** If the participant selected *health*, the strategic chatbot would present six positive and six negative personal arguments during the course of the chat. If the participant selected *environment*, the strategic chatbot would present six positive and six negative impersonal arguments. The baseline chatbot did not take the concern into account and presented three counterarguments of each type.

**DS6** After each counterargument that the chatbot presented, the participant had the choice to select *agree* or *disagree*.

**DS7** If the participant agreed, then the response was dependent on the variant of the chatbot. We implemented two slightly different variations of each chatbot (Variant I and Variant II).

  **Variant I** The chatbot presented the next counterargument.
  **Variant II** The chatbot asked *"Why do you eat meat then?"*. (This way we harvested more arguments for eating meat.)

**DS8** If the participant disagreed, the chatbot asked *"Why?"*. The participant gave an argument and depending on the length, the chatbot either asked the participant to expand or accepted it and presented the next counterargument, to which the participant agreed or disagreed and so on. The query-algorithm is explained in detail in our previous work [6].

**DS9** At the end of the chat (after presenting all 12 counterarguments and receiving a response), the chatbot would ask the participant again to select whether they *definitely wouldn't/probably wouldn't/might/probably would/definitely would* consider reducing their meat consumption.

Examples of chats with all four chatbots can be found in Appendix IV.

Table 7 Results for Variant I grouped by Baseline/Strategic and the concerns Health/Environment and their totals/averages.

| Variant I | Baseline | | | Strategic | | |
|---|---|---|---|---|---|---|
| Concern | Health | Environment | total/avg | Health | Environment | Total/Avg |
| **No of participants** | 27 | 23 | 50 | 26 | 24 | 50 |
| **Sum of intention points** | 6 | 4 | 10 (0.2) | 20 | 12 | 32 (0.64) |
| **No of harvested arguments** | 118 | 59 | 177 | 89 | 84 | 173 |
| **Avg No of disagreed CAs (out of 12)** | 4.54 | 2.56 | 3.47 | 3.42 | 3.81 | 3.46 |

*5.2.2. Results*

We divided the 50 participants for both variations of the chatbot into two groups depending on which concern they selected. For each concern group, we calculated the change of intention. Since participants were given the choice of 5 intentions (definitely wouldn't to definitely would) regarding reducing their meat consumption before and after the chat, they could either change their intention to the better, to the worse or not at all. The change in intention is the final choice of intention minus the original choice of intention. We call the units of this measure *intention points*. For example, if one participant changed her intention from "probably wouldn't" to "might" after chatting with the bot this counts as 1 intention point, whereas changing from "might" to "probably wouldn't" counts as -1 intention point. Tables 7 and 8 show the number of participants in each concern group and their average intention change within the group, as well as in total. We can see that the total number of intention points for the strategic Variant I is over 3 times higher than for the baseline of Variant I, and for the strategic Variant II the number of intention points is over 4 times higher than for the baseline Variant II.

Interestingly, the total average number of arguments that participants disagreed with while chatting with the baseline chatbot remained almost the same compared to the strategic chatbot. From this, it follows that participants do not necessarily disagree with counterarguments that do not address their concern. But despite that, those counterarguments do not have an impact on their intention. This may imply that concern-induced relevance could, for instance, be used similarly to preferences or relation weights to measure the effectiveness of a given relation. Nevertheless, more research in that direction would be needed to obtain a clearer picture. It is not surprising that overall fewer people changed their intention to the positive when chatting with Variant II, due to its "annoying" nature. Many people were irritated by the repetitive question of "Why do you eat meat then?" after they agreed with an argument. For an example see Appendix IV [1].

Table 8 Results for Variant II grouped by Baseline/Strategic and the concerns Health/Environment and their totals/averages.

| Variant II | Baseline | | | Strategic | | |
|---|---|---|---|---|---|---|
| Concern | Health | Environment | total/avg | Health | Environment | Total/Avg |
| **No of participants** | 29 | 21 | 50 | 28 | 22 | 50 |
| **Sum of intention points** | -1 | 6 | 5 (0.1) | 12 | 10 | 22 (0.44) |
| **No of harvested arguments** | 348 | 252 | 600 | 336 | 264 | 600 |
| **Avg no of disagreed CAs (out of 12)** | 3.38 | 4.52 | 4.04 | 5.29 | 2.50 | 4.06 |

Table 9 Number of participants that changed their intention to the better and to the worse for all four chatbots.

| Participant Group | Baseline | | | | Strategic | | | |
|---|---|---|---|---|---|---|---|---|
| Concern | Health | | Environment | | Health | | Environment | |
| Change of intention | Worse | Better | Worse | Better | Worse | Better | Worse | Better |
| No of participants in Variant I that changed intention | 1 | 5 | 2 | 7 | 0 | 17 | 0 | 11 |
| No of participants in Variant II that changed intention | 4 | 3 | 3 | 5 | 0 | 10 | 0 | 12 |

Tables 9 shows how many participants changed their intention to the worse and to the better, disregarding the number of intention points. We can see that in the baseline chatbots there were participants who reacted negatively to the chat and changed their intention to the worse, whereas none of the participants changed their intention to the worse after chatting with the strategic chatbots. We used the number of participants who changed their intention to the better in order to calculate the statistical significance of the difference between the control group that chatted with the baseline chatbot and the group that chatted with the strategic chatbot. Table 10 shows the p-values for both chatbots and each value group, using Chi-Square test.

All results are statistically significant apart from the environment group for Variant I because the increase in the number of participants who changed their intention to the better was not very high (from 7 in the control to 11 in the strategic group). However, out of the 7 participants who changed their intention to the better in the control group, 5 changed from "probably would" to "definitely would", which implies that they were already considering reducing their meat consumption and therefore more receptive of the presented counterarguments which resulted in the change of one intention point. In the strategic environment group, however, only 2 out of the 11 changed from "probably would" to "definitely would", whereas the rest changed from "probably wouldn't" to "might" and from "might" to "probably would". It could be argued that this is harder to achieve since the participants were less likely to consider reducing their meat consumption. These results support our hypothesis H3, that presenting arguments that address the user's concern is more likely to have a positive impact on changing the user's attitude, than presenting arguments that ignore the user's concern. All the chat data can be found in Appendix V [1].

Table 10 Statistical significance between baseline and strategic chatbots for Variant I and Variant II in total (both values) and for each value group (Health/Environment)

| Chatbot | Variant I | | | Variant II | | |
|---|---|---|---|---|---|---|
| Concern | Health | Environment | Both | Health | Environment | Both |
| p-value Chi-Square | <0.001 | 0.278 | 0.001 | 0.022 | 0.039 | 0.002 |

## 6. Discussion

Our contribution in this paper is fourfold. Firstly, we have shown that some types of arguments are considered more appealing than others in the behaviour change domain. Direct counterarguments and suggestions were the least popular in our study. Suggestions might not necessarily be unappealing but simply not tailored to the specific argument of the persuadee and therefore not relevant. Direct counterarguments, on the other hand, might trigger negative feelings from the persuadee. As pointed out in [14], focusing purely on the correctness of the arguments has led to the development of software for persuasion which, despite being correct in the dialectical argumentation sense, has been considered offensive and judgmental by its users. For the remaining argument types, there was no considerable difference in their appeal. In a user study on the persuasiveness of healthy eating messages [21], positively framed messages were shown to be more persuasive than negatively framed messages. We did not measure persuasiveness, but in our study, positive and negative consequences were equally appealing.

Secondly, we have shown that people prefer arguments that address the concern that they perceive as more important. This is not surprising, however, *concerns* are often ignored when judging the effectiveness of arguments or choosing a strategy. There are some studies that make use of different personality traits of the user attributes in order to evaluate what sort of argument might be more effective for this particular person (for examples see [13,9,23,19]). However, computational argumentation largely focuses on sentimental [7], rhetorical [11] and structural [4] attributes of the argument, rather than attributes about the user. See [20] for a review of strategies in multi-agent argumentation.

We have shown that without knowing anything about the personality of the user, and by simply asking them what they are more concerned about, we can present arguments that have a positive impact on their intention to change their behaviour. This leads to our third contribution. We have shown that presenting arguments that address the user's concern is more likely to have a positive impact on changing the user's attitude, than presenting arguments that ignore the user's concern.

And lastly, we considered how a chatbot with no natural language understanding can engage in an argumentation dialogue and influence the user's attitude towards a certain topic. Chatbots that engage in health promotion and behaviour change have recently gained interest in industry and academia. There are chatbots that encourage you to go to the gym more often like *Atlas* (`www.facebook.com/getatlasfit/`), track your mood in order to make you feel better like *Woebot* [10] (`https://woebot.io/`) and give you nutrition tips, like *Forksy* (`https://getforksy.com/`). None of these chatbots, however, use argumentation as a key component. They use a combination of reminders, provision of information and games (challenges like doing certain exercises or cooking specific meals). Our approach of crowdsourcing some of the main arguments on why people engage in a certain behaviour and the corresponding counterarguments of various types that are then used by a chatbot to engage in persuasion dialogues is a novel approach in the behaviour change domain. Using crowdsourced arguments does not require professional research but solely relies on the input of participants described in steps A1 to A3. Using only the highest ranked counterarguments assures that no inappropriate arguments are chosen for the chatbot. There are, however, also potential risks to consider. For example the spread of invalid arguments which, despite being popular, might contain wrong information.

In the future, we want to explore more argument types and their suitability and popularity in the behaviour change domain. Further, we want to use our chatbot in a different domain and research how *concerns* can be acquired for different domains. We also want to explore potential risks of our approach like the possibility to use factually inaccurate arguments.